\documentclass[10pt,twocolumn,letterpaper]{article}

\usepackage{iccv}
\usepackage{times}
\usepackage{epsfig}
\usepackage{graphicx}
\usepackage{amsmath}
\usepackage{amssymb}
\usepackage{xcolor}

\usepackage{empheq}
\usepackage{bm}
\usepackage{centernot}
\usepackage{soul}
\usepackage{stmaryrd}
\usepackage{algorithm2e}
\usepackage{framed}
\usepackage{pifont}
\usepackage{subfigure}
\usepackage[subtle, title=normal, bibliography=normal]{savetrees}
\usepackage{nicefrac}

\newcommand{\bb}[1]{\textbf{#1}}
\newcommand{\mbb}[1]{\mathbb{#1}}
\newcommand{\mc}[1]{\mathcal{#1}}

\usepackage{cancel}

\newtheorem{mydef}{Definition}

\usepackage[pagebackref=true,breaklinks=true,letterpaper=true,colorlinks,bookmarks=false]{hyperref}

\iccvfinalcopy % *** Uncomment this line for the final submission

\def\iccvPaperID{2632} % *** Enter the ICCV Paper ID here

\ificcvfinal\fi
\begin{document}

\newcommand{\cmark}{\color{green}\ding{51}}
\newcommand{\xmark}{\color{red}\ding{55}}

%%%%%%%%% TITLE
\title{Interpretable Transformations with Encoder-Decoder Networks}

% \author{Daniel E. Worrall\\
% %{\tt\small d.worrall@cs.ucl.ac.uk}
% \and
% Stephan J. Garbin\\
% %{\tt\small s.garbin@cs.ucl.ac.uk}
% \and
% Daniyar Turmukhambetov\\
% %{\tt\small d.turmukhambetov@cs.ucl.ac.uk}
% \and
% Gabriel J. Brostow\\
% %{\tt\small g.brostow@cs.ucl.ac.uk}\\
% University College London
% }

\author{
Daniel E.~Worrall \qquad Stephan J.~Garbin \qquad Daniyar~Turmukhambetov \qquad Gabriel J.~Brostow\\
%{\tt \small \{{d.worrall}, {s.garbin}, {d.turmukhambetov}, {g.brostow}\}{@cs.ucl.ac.uk}}\\
University College London \thanks{\texttt{http://visual.cs.ucl.ac.uk/pubs/interpTransform/}}
}

\maketitle
%\thispagestyle{empty}

%%%%%%%%% ABSTRACT
\begin{abstract}
Deep feature spaces have the capacity to encode complex transformations of their input data. However, understanding the relative feature-space relationship between two transformed encoded images is difficult. For instance, what is the relative feature space relationship between two rotated images? What is decoded when we interpolate in feature space? Ideally, we want to disentangle confounding factors, such as pose, appearance, and illumination, from object identity. Disentangling these is difficult because they interact in very nonlinear ways. We propose a simple method to construct a deep feature space, with explicitly disentangled representations of several known transformations. A person or algorithm can then manipulate the disentangled representation, for example, to re-render an image with explicit control over parameterized degrees of freedom. The feature space is constructed using a transforming encoder-decoder network with a custom feature transform layer, acting on the hidden representations. We demonstrate the advantages of explicit disentangling on a variety of datasets and transformations, and as an aid for traditional tasks, such as classification.
\end{abstract}

%%%%%%%%% BODY TEXT
%Our main contributions
%\begin{itemize}
%	\item Easily interpretable transformation properties---disentangling
%    \item Equivariance in fully-connected layers
%    \item Invariance for free
%    \item transformer layer
%    \item New method to learn equivariance through reconstruction
%\end{itemize}

\section{Introduction}
\begin{figure}
	\includegraphics[width=\linewidth]{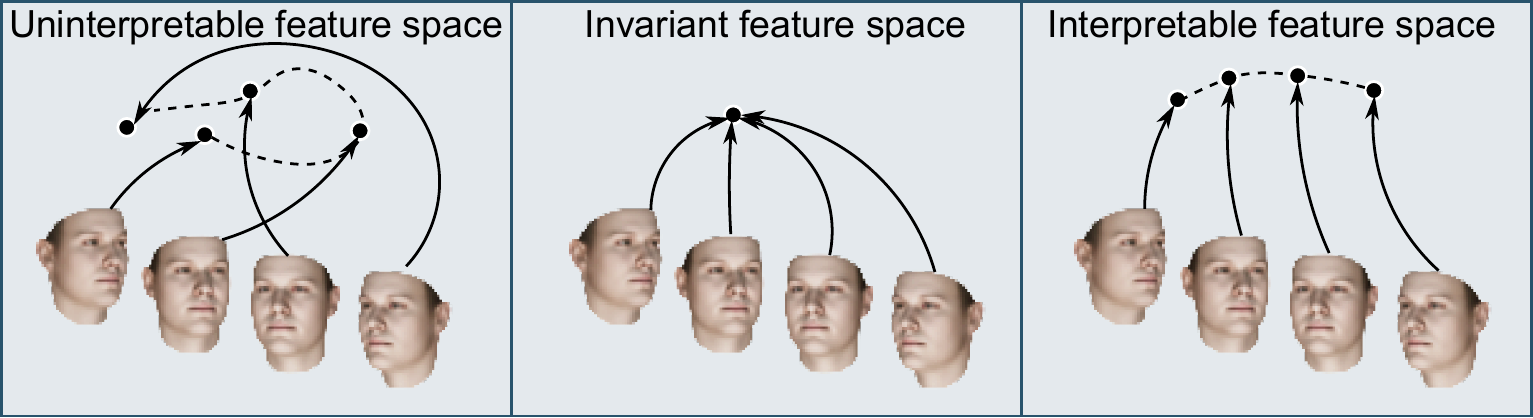}
    \caption{Three alternative feature spaces and how each encodes images of the same person. (Left) A feature space that is hard to interpret, similar to one learned by a typical CNN. While transformation information is present, it is not obvious how to extract that directly from the feature space. (Middle) A transformation-\emph{invariant} feature space. (Right) An interpretable feature-space, where ordered transformations of the input subject relate to ordered, structured features. This is like a learned metric space, but also allows for image synthesis. Images of another person are not shown, but would ideally project similarly, albeit elsewhere in each feature space.}
    \label{fig:feature_spaces}
    \vspace{-1em}
\end{figure}

We %consider one aspect of deep feature-spaces, namely how 
seek to understand and exploit the deep feature-space relationship between images and their transformed versions. Different feature spaces are illustrated in Figure~\ref{fig:feature_spaces}, and support different use-cases: separability helps discriminate between categories such as identity, while invariance improves robustness to nuisance variables during data capture. Taking head pose as an example, what is a nuisance for one task could be the focus of another. Therefore, we propose deep features with transformation-specific \emph{interpretability}, which combine both (1) discriminative and (2) robustness properties, with %makes categories of images separable, enables robustness to nuisance variables, but further,  
the further benefits of (3) a user-guided parameterized space for controlling image synthesis through interpolation. 
%\cut{Visualizing the structure of deep feature spaces is now a key problem and opportunity for computer vision. } 

% This problem 
Learning such a feature space is difficult. In image data, transformations of objects usually couple in complex nonlinear ways, leading to an \emph{entangling} of transformations. The reverse process of \emph{disentangling} is then especially hard. An obvious post hoc solution is to learn disentangling transformations using a regressor~\cite{Lenc15}, but this is a time-consuming and inexact process. We cannot assume that the change in representation of a chair and its rotated twin is necessarily the same as the change in representation between a banana and its equally rotated twin. 
%We present a method to learn representations that transform according to predefined rules.
We propose disentangling as an end-to-end supervised learning problem. Some image variations are hard to quantify or explain. But others, for instance 2D and 3D warps or color appearance changes, allow ready access to pre- and post-warp image pairs, along with their ground-truth transformation parameters. These easier transformations, we find, lend themselves to smooth parameterization in feature space, and therefore interpretability. One could argue that it is nicer to learn everything only from raw data, but the transformation parameter labels considered here are obtained with little or no human effort. We therefore pre-define the feature-space structures that encode basic transformations, and train neural networks that map into and out of this feature-space.

We take our motivation from considering the feature space structure, introduced by convolutional neural networks \cite{LeCun89} (CNNs). %CNNs are proving successful in many traditional computer vision tasks. 
CNNs owe their success to two differences from the older and more general multilayer perceptrons~\cite{lippmann1987introduction}: 1) the receptive field of deep neurons is localized to a small neighborhood, typically not greater than $7\times 7$ pixels from the layer below, and 2) incoming weights are tied between all translated neurons. The motivation behind translational weight-tying is that correlations in the activations are invariant under translation. The side-effect of enforcing such a structure on the weights of a neural network is that integer pixel translations of the image input induce proportional integer pixel translations of the deep feature maps. This phenomenon is called \emph{equivariance}, meaning the feature-representation of a shifted input is the same, save for its location. We explore continuous transformation equivariance for CNNs, and for the first time, for fully connected models. % is undistorted, while reflecting shifts of the input image. 

In this paper, we consider rotations in 2D and 3D, out-of-plane rotations, small translations, stretchings, uniform scalings and changes in lighting direction. For these transformations CNNs do not generally display the equivariance property; although, there are a number of works, which do tackle the problem of rotation \cite{Cohen16,Dieleman16,Oyallon15,Fasel06,Gonzalez16,Laptev16,Worrall16,Gonzalez16a,Zhou17}. The main problem with all these approaches (which we detail in the next section) is that the equivariance properties are handcrafted, and suffer from unmodeled oversights in the design process. For instance, all but \cite{Worrall16} consider equivariance to discretely sampled rotations, when real world rotations are in fact continuous. Given that we can simulate many image-space transformations, it seems only natural to simply acquire equivariance through learning.

We now cover related work and theory, followed by Section~\ref{sec:method} where we introduce our method and the new \emph{feature transform layer}, and Section~\ref{sec:experiments} where we test our framework on de-render--re-render problems and for view independent features.

\section{Related Work and Theory}
\label{sec:related_work}
Here we outline basic concepts for us to formalize the task of encoding interpretable transformations, and break down a list of related works into categories of handcrafted or learned equivariance in traditional vision and deep learning.

{\mydef \label{def_equi}
A function {\normalfont $\bb{f}:\mathcal{X}\to\mathcal{Y}$} is \emph{equivariant} \cite{Wilson88} under a set of transformations $\Theta$ if for any transformation $\mc{T}:\Theta\times\mc{X}\to\mc{X}$ of the input, we can associate a transformation $\mc{F}:\Theta\times\mc{Y}\to\mc{Y}$ of the output such that
{\normalfont
\begin{align}
	\mc{F}_{\theta}[\bb{f}(\bb{x})] = \bb{f}(\mc{T}_{\theta}[\bb{x}]), \label{eq:equivariance}
\end{align}
}
for all $\theta\in \Theta$. Transformations $\mc{T}_{\theta}$ and $\mc{F}_{\theta}$ represent the same underlying transformation but in different spaces, denoted $\theta$.} 

Equivariance is desirable, because it reveals to us a direct relationship between image-space and feature-space transformations, which for deep neural networks are usually elusive \cite{Lenc15}. Note that \emph{invariance} is a special case of equivariance, where $\mc{F}_{\theta}=\mbb{I}$ is the identity for all input transformations.

{\mydef \label{def_interp}
We define an \emph{interpretably equivariant feature-space} to be an equivariant feature-space as in Equation \ref{eq:equivariance}, where the transformation functions $\mc{F}_{\theta}$ and $\mc{T}_{\theta}$ are quantitatively known and can be implemented for all $\theta$, {\normalfont $\bb{x}$} and {\normalfont $\bb{f}$}.
}

At an abstract level, an equivariant function is one where some level of structure is preserved between the input and output. Interpretability is the added requirement that for a given $\theta$ we know how to apply $\mc{F}_{\theta}$ and $\mc{T}_{\theta}$. It may be the case that one of these transformations is complicated and cannot be written down as a mathematical expression in closed form (e.g., the rendering equation), but as long we are able to simulate it that is enough. As we show in Section \ref{sec:feature_transform_layer}, one way of preserving the structure of transformations across a feature mapping is via a condition called the \emph{homomorphism property}. In all of the subsequent related works, equivariance to transformations is the central theme.

\textbf{Handcrafted methods}
In the 1980s, Crowley and Parker \cite{Crowley84} studied scale-space representations. These are formed by convolving images with scaled versions of a filter. Scale-space methods exhibit interpretable equivariance. They can be extended to invertible transformations by transforming the filters \cite{Lindeberg11,Alcantarilla12} but has computational complexity exponential in the number of degrees of freedom (DOF) of the transformation. Furthermore, we can only convolve with a finite number of filters, when in reality many transformations are continuous. Freeman and Adelson \cite{FreemanAdelson91} and Lenz \cite{Lenz90} simultaneously solved the continuity problem, through orientation steerable filters $w_\theta$. These can be synthesized at any continuous orientation $\theta$. These are formed as a linear combination of fixed basis filters $\phi_n$:
\begin{align}
	w_\theta(\bb{x}) = \sum_{n=1}^N \alpha_n(\theta) \phi(\bb{x}). \label{eq:steerable_filters}
\end{align}
$\alpha_n(\theta)$ are known as the \emph{interpolation functions}. These are still band-limited but unlike scale-space the frequency characteristics are easier to design.
%These are especially useful when we wish to explore the convolutional response of an image to a filter at orientation $\theta$, since
%\begin{align}
%	[w*I](\bb{x},\theta) = \sum_{n=1}^N \alpha_n(\theta) [\phi_n*I](\bb{x}),
%\end{align}
%where $[w*I](\bb{x},\theta)$ is the response for a filter $\theta$-rotated about pixel $\bb{x}$. We see the orientation response can be probed at any continuous rotation after the convolution has taken place with just $O(N)$ add--multiplies. The response is continuously equivariant in $\theta$.
Steerable filters were extended to most transformations with one DOF (one-parameter subgroups) \cite{Teo98,Simoncelli92}, for instance, 1D translations, 2D rotations, scalings, shears, and stretches. For these transformations, there is a function $\rho$, under which transformation $\theta$ becomes a shift, so $I(x)\overset{\mc{T}_\theta}{\to} I(\rho^{-1}(\rho(x)-t_{\theta}))$, where $t_{\theta}$ is the shift. Meanwhile, Perona \cite{Perona91} showed that in practical situations some transformations cannot be enacted exactly using steerable functions, for instance scale and affine transformations (specifically those which do not have compact group structure). He showed these can be approximated well with very few basis functions, computed from the singular value decomposition of a matrix of transformed versions of a template patch. This is limited by template choice, SVD efficiency, and figuring out the interpolation functions for steering. More recently Hasegawa \cite{Hasegawa15} and Koutaki \cite{Koutaki14} used a variant of this method to learn an affine-equivariant feature detector.

Invariance to 1 DOF transformations can be gained via the Fourier Transform (FT) Modulus method \cite{Kokkinos12}. This uses the time-shifting property of the FT $w(x-t) \overset{FT}{\iff} e^{i\omega t}\mc{W}(\omega)$, where $\mc{W}(\omega)$ is the FT of $w(x)$. The FT modulus $|e^{i\omega t}\mc{W}(\omega)|=|\mc{W}(\omega)|$ is independent of the shift $t$. As noted in Scattering Networks \cite{Bruna13}, this operation removes excessive localization information and is unstable to high-frequency deformations noise. They instead take the modulus of the response to a bank of discretely rotated and scaled wavelets, repeatedly in a deep fashion. This is perhaps the most successful version of a handcrafted deep equivariant feature map.

\textbf{Neural Networks}
Equivariance in deep learning has very deep roots as far back as the early 1990s. Barnard and Casasent \cite{Barnard91} split the main approaches to transformation invariance into three categories: 1) \emph{Data augmentation}: This is effective and simple to implement, but lacks interpretability. 2) \emph{Preprocessing}: This is effective, but cannot be applied to geometric transformations. 3) \emph{Structured weight networks}: These are numerous in the literature. CNNs \cite{LeCun89} are the most famous example. Pixel-wise integer shifts of an input image will induce proportional pixel-wise shifts in the deep feature space. For partial translation invariance, there is the Global Average Pooling layer \cite{Lin13}. For rotations there are two major approaches for discrete rotations: rotate the filters \cite{Cohen16,Cohen16a,Gonzalez16,Oyallon15,Gonzalez16a,Zhou17} and rotate the input/feature maps \cite{Dieleman16,Fasel06,Laptev16}. Continuous rotations were recently proposed by \cite{Worrall16}. They restrict their filters and architectures so that the convolutional response is equivariant to continuously rotated inputs. Beyond rotation, \cite{Henriques16} warp the input, so that general transformations are globally linearized, facilitating the application of CNNs. This requires prior knowledge of the type of transformation and where it is applied in the image. \cite{Cohen16a} can deal with multiple transformations, but these are restricted to group-theoretic structures. \cite{Jaderberg15} are able to explicitly transform feature maps with the spatial transformer layer, but do not transform features in the channel dimension. In contrast to the above methods, our method is general and does not require extensive architectural engineering. We can also disentangle confounding factors such as out-of-plane rotation and lighting direction.

\textbf{Deeply Learned Equivariance}
Some have sought to learn equivariance directly from data. These broadly split into purely generative, purely discriminative and auto-encoded methods. \textbf{Discriminative}:\ \cite{Lenc16} regress affine equivariant feature-descriptors directly using supervised data. Their framework is easy to implement, but restricted to group-theoretic transformations. \textbf{Generative}:\ \cite{dosovitskiy15} generate views of 3D chairs by regressing appearance with a CNN from an embedding space. In InfoGAN, \cite{Chen16} instead used a mutual information maximizing criterion for unsupervised learning of the `natural' transformations in a training set. This mostly manages to disentangle transformation, but unlike \cite{dosovitskiy15} is non-interpretable. \textbf{Auto-encoded}: \cite{Kulkarni15} presented the deep convolutional inverse graphics network (DC-IGN), a partially supervised variational auto-encoder \cite{Kingma13}, equivariant to out-of-plane rotation and relighting. Their model is impressive but requires a complicated training procedure, is partially interpretable, and unlike us does not fully exploit known supervised information about transformations. \cite{Memisevic10,Hinton11,Zhou16} instead reconstruct transformed versions of an image, given the image and transformation parameters as input. These are similar to our method, but cannot be used to extract interpretable transformation equivariants, which we can do. \cite{cohen2015representations} does learn interpretable equivariance to manipulate images of 3D objects from 2D images, but this is only demonstrated on 3D rotations. \cite{rezende2016} also does learn interpretable equivariance for 3D volumes from 2D images, but their representation space is entire 3D volumes. This is impressive, but it is computationally expensive to represent entire volumes in memory, when sometimes it may not be necessary.

\section{Method}
\label{sec:method}
CNNs are interpretably equivariant to pixel-wise translations of their input up to boundary effects, but not to transformations such as 2D and out-of-plane rotations, uniform scalings, stretches, relighting, flips, etc. In this section we design a neural network to learn an interpretable transformation equivariant feature-space. Our method can cope with continuous transformations on intervals, for example, uniform scalings and stretches, and continuous transformations on circles, such as, geometric rotation and relighting, but not discrete transformations, like vertical flips. In Section \ref{sec:problem_setup} we outline our general framework and in Section \ref{sec:feature_transform_layer} we introduce the \emph{feature transform layer}, a channel-wise analogue of the spatial transformer, which can also be applied to fully-connected layers.

\subsection{Problem Setup}
\label{sec:problem_setup}
We assume that we are given a training set $\mc{D} = \{(\bb{x}^1,\tilde{\bb{x}}_{\theta^i}^1,\theta^1), ..., (\bb{x}^N,\tilde{\bb{x}}_{\theta^i}^N,\theta^N)\}$ containing pairs of views of transformed examples $(\bb{x}^i,\tilde{\bb{x}}_{\theta^i}^i)$ and relative transformation vectors $\theta^i$. The relative transformations may be the result of a sensor measurement, or they may be the result of artificial data augmentation, in which case the training set is potentially infinite. The task is to predict $\tilde{\bb{x}}_{\theta^i}^i$ given $\bb{x}^i$ and $\theta^i$ (from now on we just write $\theta$ for short). We use relative transformation information instead of absolute transformations, because there is no canonical pose, which generalizes across object classes, where alignment between, say, a banana and an airplane does not make sense.

%{\color{gray}We consider two major groups of transformations: continuous geometric transformations and continuous appearance changes. For both, the domain $\Theta$ of the transformation parameters $\theta$ is a smooth manifold, so transformations such as flips and mirroring are not supported. Geometric transformations change the spatial relationship between the camera and the observed objects. Appearance transformations affect the individual pixel values for points on the object.}

Many images $\bb{x}\in\mc{X}$ are formed from capturing an object $\bb{o}\in\mc{O}$ in the 3D world projected via a function $\Pi:\mc{O}\to\mc{X}$ onto a 2D canvas. To transform image $\bb{x}$ into $\tilde{\bb{x}}_{\theta}$ we have to invert $\Pi$ to find $\bb{o}$, perform the world-space transformation and re-project back into image space, so
\begin{align}
	\tilde{\bb{x}}_{\theta} = \Pi \left [\mc{T}_{\theta}[\bb{o}] \right ] = \Pi \left [\mc{T}_{\theta} \left [\Pi^{-1}[\bb{x}]\right ] \right ].
\end{align}
The problem with this approach is that $\Pi$ is in usually non-invertible.
\begin{figure}
	\includegraphics[width=\linewidth]{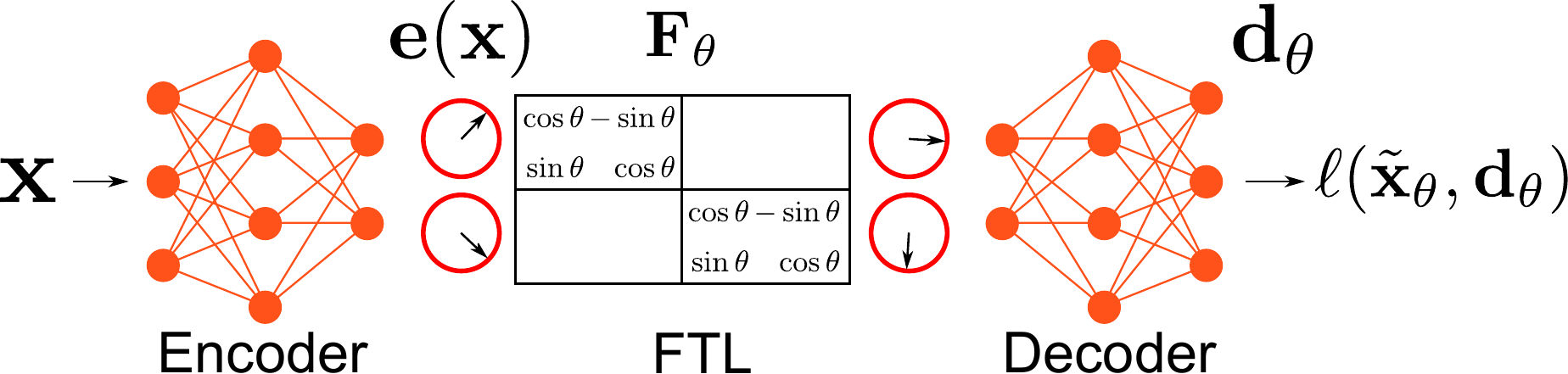}
    \caption{We enforce equivariance by minimizing the loss $\ell$ between reconstruction of transformed features $\bb{d}_{\theta}$ and a transformed target $\tilde{\bb{x}}_{\theta}$. Given just $\bb{x}$, the encoder-decoder network does not have enough information to produce a transformed output, thus supplying the missing information $\theta$ via the feature transform layer (FTL) forces the network to learn a mapping in and out of the FTL. Critically, whereas other approaches, such as transforming auto-encoders \cite{Hinton11} and InfoGAN \cite{Chen16}, learn the reconstruction to be sensitive to feature transformation information, we can simultaneously learn to map from images to transformation equivariant features.}
    \label{fig:system}
\end{figure}
Our solution is to infer the 3D object $\bb{o}$ given $\bb{x}$ via statistical methods. CNNs are good at this kind of task (e.g.,\ \cite{Kulkarni15}), so we opt to use a CNN. Now storing a full volumetric representation like in \cite{rezende2016} is costly, so we instead opt to use a compressed feature encoding $\bb{e}(\bb{x})$ to approximately represent $\bb{o}$, this requires we also have a feature-space representation of the transformation, $\mc{F}_{\theta}$---see Section \ref{sec:feature_transform_layer} for details. In our case the feature space is partially learnable, with pre-defined structure imposed by $\mc{F}_{\theta}$. Our basic model is shown in Figure \ref{fig:system}, it is an encoder-decoder network. Loosely speaking 
\begin{align}
	\bb{e}(\bullet) 	&\text{ approximates } \Pi^{-1}[\bullet], \notag \\
        \mc{F}_{\theta}	&\text{ is the feature space equivalent to } \mc{T}_{\theta}, \notag  \\
        \bb{d}(\bullet) 	&\text{ approximates } \Pi[\bullet], \notag 
\end{align}
where we have written $\Pi^{-1}[\bullet]$ to mean inversion of the projection if possible, or approximation of it. We train the weights of the encoder and decoder by minimizing a summed reconstruction loss $\ell$, where
\begin{align}
	\mc{L}(\mc{D}) = \sum_{i} \ell \left (\bb{d} \left (\mc{F}_{\theta^i} \left [\bb{e}(\bb{x}^i) \right ] \right ), \tilde{\bb{x}}_\theta^i \right ).
\end{align}
In our experiments we use a diverse set of losses, namely, L1 loss, SSIM, and balanced cross-entropy. Note that since we define $\mc{F}_{\theta}$ the feature space of encodings $\bb{e}(\bb{x})$ is interpretable by Definition \ref{def_interp}. In Section \ref{sec:feature_transform_layer}, we demonstrate an encoding, which enforces explicit disentangling and from which we can gain approximate transformation invariance `for free'.

\begin{table*}[t]
  \begin{center}
    \begin{tabular}{|l|c|c|c|c|c|c|c|c|}
    \hline
    Method & $\tilde{\bb{x}}_{\theta} | \theta$ & $\tilde{\bb{x}}_{\theta} | \theta, \bb{x}$ & $\theta | \bb{x}$ & CNN & MLP	& Interpretable	& Supervised & Image size\\
    \hline
    DC-IGN \cite{Kulkarni15}				& \cmark	&	\xmark	& * & \cmark	& \cmark & \dag 		& \ddag 	& 150x150 \\
    InfoGAN \cite{Chen16} 					& \cmark 	&	\xmark	& \xmark & \cmark	& \cmark & \xmark 	& \xmark 	& 64x64  \\
    Generating Chairs \cite{dosovitskiy15}	& \cmark 	&	\xmark	& \xmark & \cmark	& \cmark & \cmark 	& \cmark 	& 128x128\\
    Transforming AEs \cite{Hinton11} 		& \xmark	& \cmark & \xmark & \xmark	& \cmark & \xmark 	& \cmark 	& 96x96\\
    Learned Visual Reps. \cite{cohen2015representations}& \xmark  & \cmark& \xmark 	& \xmark & \cmark 	& \cmark & \xmark & 96x96\\
    Unsup. 3D from images \cite{rezende2016}& \xmark 	& \cmark & \xmark & \cmark 	& \cmark & \cmark 	& \xmark	& 30x30x30 \\
    Covariant features \cite{Lenc16}		& \xmark	& \xmark& \cmark & \cmark	& \cmark & \cmark 	& \cmark 	& 57x57\\
    Spatial Transformer \cite{Jaderberg15} 	& \xmark			& \cmark & \xmark	 & \cmark	& \xmark & \cmark 	& - 		& Any \\
    Ours									& \xmark 	& \cmark & \cmark & \cmark & \cmark	& \cmark	& \cmark 	& 150x150 \\
    \hline
    \end{tabular}
  \end{center}
  \caption{Comparison of method scopes. In the first 3 columns we display whether a method can generate an image $\tilde{\bb{x}}_{\theta}$ given just parameters $\theta$, $\tilde{\bb{x}}_{\theta} | \theta$; conditioned on an original image $\tilde{\bb{x}}_{\theta} | \theta, \bb{x}$; or infers transformation parameters given an image $\theta | \bb{x}$. * Qualitative relationship only. \dag Correspondence between feature dimensions and transformations known, qualitative relationship only. \ddag Partial supervision: minibatches grouped into variation of single parameter, but values not given.}
  \label{tab:comparisons}
\end{table*}

\subsection{The Feature Transform Layer}
\label{sec:feature_transform_layer}
The feature-space equivalent of the image-space transform $\mc{T}_{\theta}$ is the \emph{feature transform layer} $\mc{F}_{\theta}$. It is an analogue of the spatial transformer \cite{Jaderberg15}, but applied to general feature-spaces, not necessarily with spatial dimensions. This means that we can apply it to fully connected layers as well as convolutional layers. It is easiest to describe the feature transform layer via its implementation.

Consider a feature vector $\bb{e}$, which may be a column of CNN feature channels above a pixel location in an image, or the output of a fully-connected layer. The feature transform layer performs a linear transformation of $\bb{e}$ via matrix $\bb{F}_\theta$, such that the output $\bb{y}$ of the layer is
\begin{align}
	\bb{y} = \mc{F}_\theta[\bb{e}] = \bb{F}_\theta \bb{e}.
\end{align}
We only consider linear transformations, where
\begin{align}
	\bb{F}_{\theta_2\theta_1} = \bb{F}_{\theta_2}\bb{F}_{\theta_1}. \label{eq:homomorphism}
\end{align}
This condition says that if we apply transformation $\theta_1$ to an image, followed by transformation $\theta_2$, which we have written as $\theta_2\theta_1$, then in feature space this should be equivalent to applying $\bb{F}_{\theta_1}$ followed by $\bb{F}_{\theta_2}$. We refer to Equation \ref{eq:homomorphism} as the \emph{homomorphism property}. Abstractly, we can think about it as forcing the neural network to learn a mapping from image-space to feature-space, which preserves the intrinsic structure of the transformations. The homomorphism property implies that (see Supplementary Material)
\begin{align}
	\bb{F}_{\theta_1^{-1}} = \bb{F}_{\theta_1}^{-1}.
\end{align}
This means that invertible transformations of the input are invertible in feature-space. The homomorphism property is key to ensuring that transformation information is not lost when mapping into feature-space. Examples of $\bb{F}_{\theta}$ are N-dimensional rotation matrices, also known as $SO(N)$, full-rank diagonal matrices, or most generally the group of invertible $N\times N$ matrices, known as $GL(N)$.
\begin{figure}[b]
\vspace{-1em}
\begin{center}
	\includegraphics[width=0.65\linewidth]{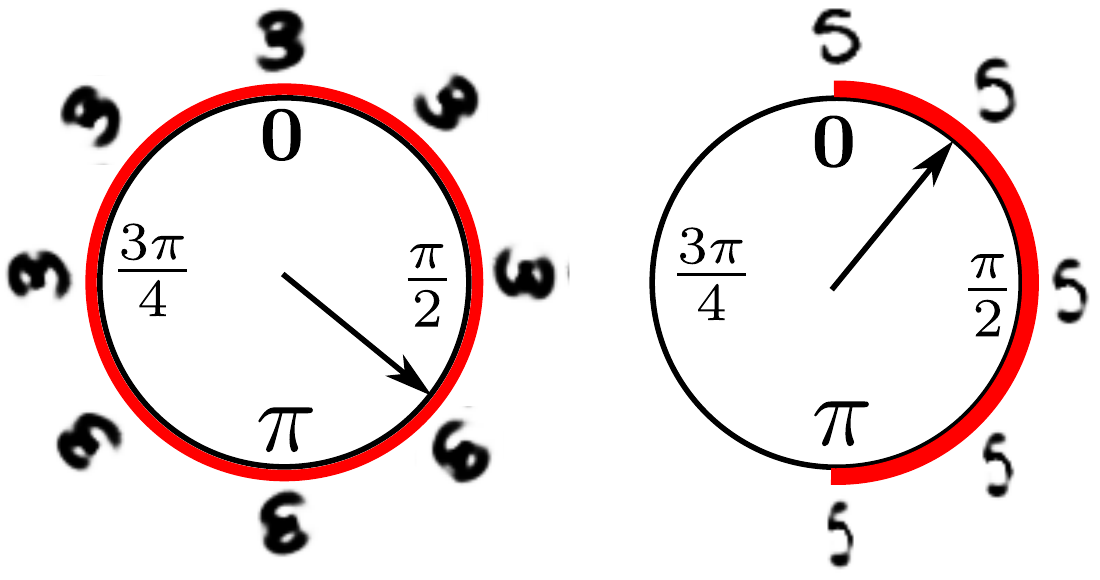}
    \caption{We encode transformations by mapping them on to circles and N-dimensional hyperspheres in feature space. This parameterization can deal with periodic and bounded transformations on an interval. The L2-norm of the result feature vectors are transformation invariant.}
    \label{fig:circles}
\end{center}
\vspace{-1em}
\end{figure}
We use rotation matrices, $\bb{R}_\theta$, which have the additional property of being orthogonal or \emph{norm-preserving}. This means that we can use the feature vector lengths as transformation invariants because
\begin{align}
	\|\bb{R}_\theta\bb{e}\|_2^2 = \bb{e}^\top\bb{R}_\theta^\top\bb{R}_\theta\bb{e} = \bb{e}^\top\bb{e} = \|\bb{e}\|_2^2, \label{eq:norm}
\end{align}
which shows that $\|\bb{R}_{\theta}\bb{e}\|_2^2$ is in fact independent of $\theta$. Feature vectors are usually high-dimensional consisting of many channels. We thereform implement the feature transform layer by applying the same rotation matrix on multiple groupings of channels, which we call \emph{subvectors} of $\bb{e}$. We can then define a larger set of invariants, by measuring the relative phase between different subvectors. These are invariant to $\theta$, because if $\bb{e}_1$ and $\bb{e}_2$ are two subvectors of $\bb{e}$, then
\begin{align}
	(\bb{R}_{\theta}\bb{e}_2)^{\top}\bb{R}_{\theta}\bb{e}_1 = \bb{e}_2^{\top}\bb{R}_{\theta}^{\top}\bb{R}_{\theta}\bb{e}_1 = \bb{e}_2^{\top}\bb{e}_1, \label{eq:phase}
\end{align}
which is independent of $\theta$. If $\bb{e}_1=\bb{e}_2$, this reduces down to the feature vector length. We denote the concatenation of all subvector dot products as $\|\bb{e}\|_\mc{F}$. While at first not obvious, we can encode many transformations using rotation matrices, even ones which do not have periodic structure. The trick is to map the domain of the transformation onto the half-circle/sphere, see Figure \ref{fig:circles}. We prefer to do this rather than using another, perhaps more natural, representation because of the convenience of taking L2-norms and inner products to form invariants.

\textbf{Disentangling}
We now consider how to disentangle transformations. Since we can model transformations, whose parameters exist on a circle or interval, we can model each independent transformation DOF by mapping it to a different circle or half-circle. Some transformations, like lighting direction, are more conveniently mapped to the surface of a 3D-sphere. Thus the feature transform layer is
\begin{align}
	\bb{F}_{\theta}\bb{e} = \begin{bmatrix}
	\bb{R}_{\theta_1} &			& \\
    				& \ddots	& \\
    				&			& \bb{R}_{\theta_N} \\
	\end{bmatrix} \bb{e},
\end{align}
with possible tied $\theta_i$ when we apply a transformation to multiple subvectors. The feature transform layer is simple to implement---it is just a matrix multiplication and the block diagonal structure allows efficiency saving via reshapes. In our experiments we found a slow down of just 2\%. Furthermore, it can be applied to convolutional features in synchrony with a spatial transformer \cite{Jaderberg15} for complete control of both spatial and feature properties.

\section{Experiments, Results, and Discussion}
\label{sec:experiments}d
Below we demonstrate the ability of our system to learn meaningful features on MNIST \cite{mnist}, MNIST-\emph{rot} \cite{larochelle2007deep}, the Basel Face Dataset \cite{Paysan09}, and ModelNet10 \cite{Wu15}. We choose these datasets because they demonstrate our system's general-purpose usage and performance on 2D and 3D images, for transformations with complex entanglement, and with and without information loss. Our encoder-decoder structure is shown in Figure \ref{fig:networks}. They are all implemented in TensorFlow.

\subsection{MNIST: 2D images---2D transformations}
This experiment demonstrates our system's ability to disentangle confounding transformations and how it reconstructs an input, after manipulation of the features. The MNIST dataset \cite{mnist} contains 50k training, 10k validation, and 10k grayscale test images of handwritten digits, size $28\times 28$. The images are very simple, usually just a pen-stroke. We apply random scalings in the x- and y-directions followed by a random 2D rotation. Due to the simplicity of the images, we use an MLP for both encoder and decoder. Both encoder and decoder have 3 layers, separated by batch normalization \cite{ioffe2015bn} and leaky ReLU nonlinearities \cite{maas13} apart from the input and output of the feature transform layer, which are linear. All layers except the input and output are 510 neurons wide\footnote{We use this non-standard width because we model three transformations, with each transformation modeled on a separate circle. So feature-space dimensionality must be a multiple of $3\times 2=6$. Furthermore, the value of 510 is close to 512, a common feature-space dimensionality.}. The feature transform matrices are a block diagonal composition of three 2D rotation matrices repeated 85 times: rotation $\bb{R}_{\text{rot}}$, x-scaling $\bb{R}_{\text{scale-x}}$, and y-scaling $\bb{R}_{\text{scale-y}}$. We train with the Adam optimizer \cite{Kingma14} for 200 epochs, with minibatch size 128 and initial learning rate $10^{-3}$.
\begin{figure}[b]
    \vspace{-1em}
	\includegraphics[width=\linewidth]{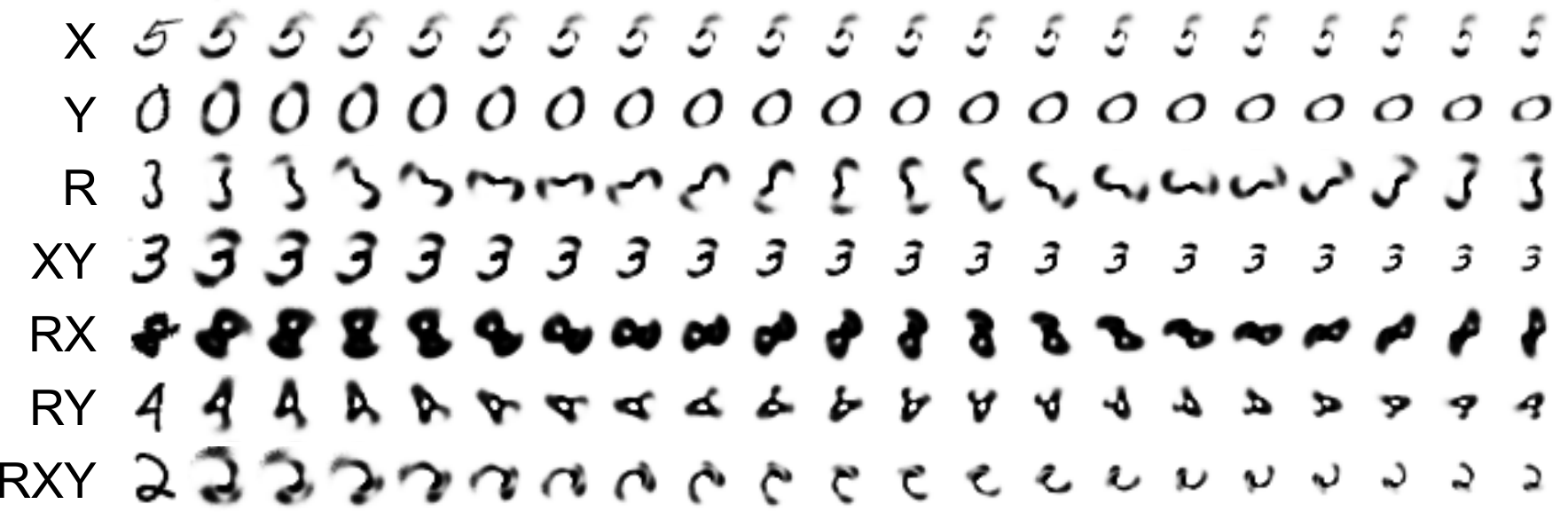}
    \caption{MNIST reconstructions: The left most column indicates transformation. The second to left column shows the input. Subsequent columns show the transformed images. The reconstruction struggles slightly with enlarged images, on the left, but on the whole clearly show that we have control over the disentangled representation. Notice that the x- and y-scalings are in the coordinate frame of the canonical pose of the digits. This demonstrates the ability to disentangle confounding transformations.}
    \label{fig:MNIST}
\end{figure}
After training we pass a random digit from the test set through the encoder and transform the code by multiplying by feature transform matrix $\bb{F}_\theta$. In Figure \ref{fig:MNIST} we show random digits from the test set, slowly varying the transformation vectors on an interval. Each row shows a random digit under a combination of rotation, x-, and y-scaling. Notice how the encoder-decoder successfully learns to rotate digits, solely from the feature transformation. Notice also that the scalings are applied in the x- and y-directions of a coordinate system aligned to the canonical pose of the digit. The system struggles when the images are magnified, nonetheless these results demonstrate clearly that we can learn a feature-space, where we have control over reconstruction transformations.
\begin{table}[b]
  \begin{center}
    \begin{tabular}{|l|c|c|}
    \hline
    Method & Test error (\%) & Flexibility \\
    \hline\hline
    SVM \cite{larochelle2007deep}				& $11.11$		& \cmark \\
    Transformation RBM \cite{sohn2012local}		& $4.2$ 		& \cmark \\
    Conv-RBM \cite{schmidt2012priors} 			& $3.98$ 		& \cmark \\
    CNN \cite{Cohen16}							& $5.03$ 		& \xmark \\
    CNN \cite{Cohen16} + data aug*				& $3.50$ 		& \xmark \\
    $P4$CNN rotation pooling \cite{Cohen16}		& $3.21$		& \xmark \\
    $P4$CNN \cite{Cohen16} 						& $2.28$		& \xmark \\
    Harmonic Networks \cite{Worrall16}			& $1.69$ 		& \xmark \\
    RotEqNet \cite{Gonzalez16a}					& $\bb{1.16}$	& \xmark \\
    \hline
    Ours (MLP variant)							& $4.90$		& \cmark \\
    Ours (CNN)									& $2.14$		& \cmark \\
    \hline
    \end{tabular}
  \end{center}
  \caption{MNIST-rot test accuracy: We beat standard CNNs with similar architectures. Even our MLP variant can beat a baseline CNN. The state-of-the-art is reserved for models specifically designed for rotation. \textsc{Flexible} models can learn general transformations, while others only deal with translation and rotation. Interestingly, we beat \cite{Cohen16}, which was designed for rotation.}
  \label{tab:MNIST_rot}
\end{table}
\textbf{MNIST-rot}
Next we explore if we could improve classification on the MNIST-\emph{rot} dataset, with an explicitly rotationally equivariant feature space. We feed the learned transformation invariant subvector relative phases $\|\bb{e}(\bb{x})\|_{\mc{F}}$ into a classifier $f$ (Figure 5) and use the output $f(\|\bb{e}(\bb{x})\|_\mc{F})$ for classification. MNIST-\emph{rot} \cite{larochelle2007deep} is a specific subset of MNIST split into 10k train, 2k validation, and 55k test images, rotated randomly on the circle. Our results are in Table \ref{tab:MNIST_rot}. While we do not achieve state-of-the-art on this benchmark, we do beat standard CNNs trained with data augmentation. All models better than us are designed specifically for rotation. This indicates that in low data scenarios, it pays to exploit our prior knowledge of how transformations affect data. We can use this knowledge to construct meaningful feature spaces, where equivariance and invariance can be utilized. We found that it helps to add a regularization term $\|\|\bb{e}(\bb{x})\|_{\mc{F}} - \|\bb{e}(\tilde{\bb{x}}_{\theta})\|_{\mc{F}}\|_2^2$ to the loss function encouraging transformed encodings of the input to be equal in length
\begin{figure}[t]
	\includegraphics[width=\linewidth]{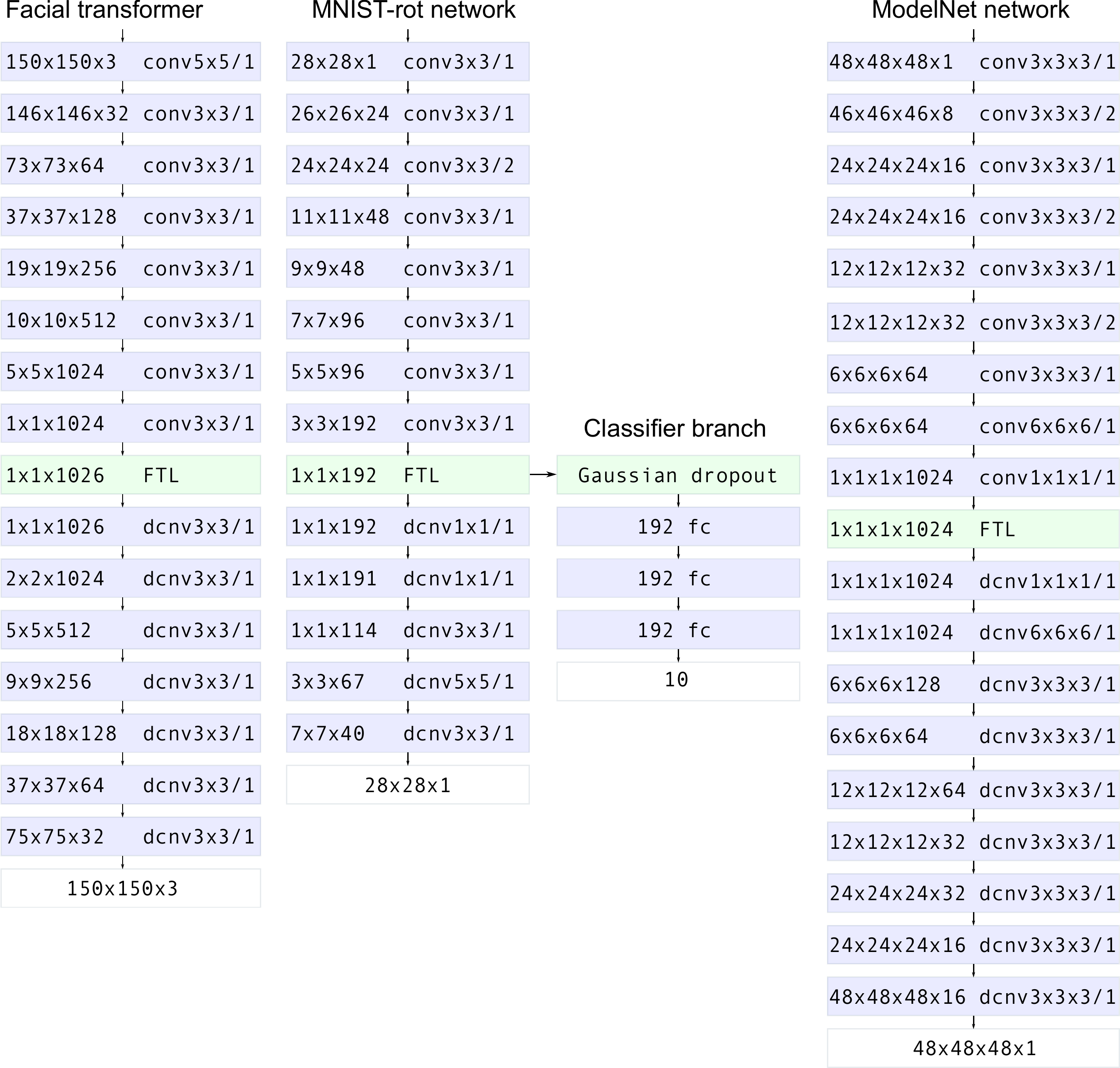}
    \caption{The architectures used in our experiments. \textsc{Left}: Facial transformer, \textsc{center} MNIST-\emph{rot} network, \textsc{right}: ModelNet network. The first set of numbers indicate input tensor shape, the second set of numbers indicate operation (conv: convolution, dcnv: deconvolution, FTL: feature transformer layer, fc: fully-connected). The number trailing the $/$ is the stride. For deconvolution, we first nearest-neighbor upsample, followed by convolution.}
    \label{fig:networks}
\end{figure}
\begin{figure*}[t]
	\includegraphics[width=\textwidth]{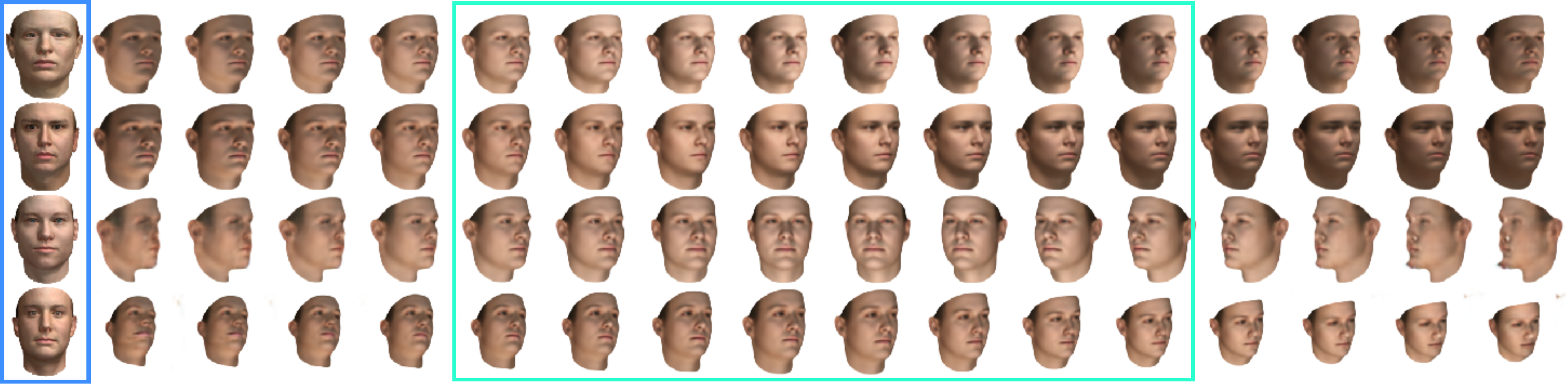}
    \caption{(Viewed best in color). Relit and re-rotated reconstructions from our Basel face encoder-decoder model. The input faces shown on the left (blue box) are not presented to the encoder-decoder at training time. From left to right we vary one degree of freedom only. Outside the large green box the encoder-decoder has never seen those transformation parameters. We note the impressive ability of the model to rotate out-of-plane and to relight a 3D surface, when only given a 2D input and a pair of 3D rotation matrices. For unseen transformation parameters, notice that the relighting is of perceptually decent quality, but that the geometric rotations degenerate in quality around the boundaries, such as the nose and chin.}
    \label{fig:faces_panel}
\end{figure*}
\subsection{Basel Faces: 2D images---3D transformations}
In this experiment, we return to disentangling transformations for superior control in reconstruction. The Basel Face dataset \cite{Paysan09} contains synthetic face renderings encoded using a PCA model. We can randomly draw faces with vertex positions $\bb{s}$ and vertex colors $\bb{t}$ from the model by sampling two 199-dimensional vectors $\bm{\alpha}$ and $\bm{\beta}$ from a unit Gaussian and retrieving the face by
\begin{align}
	\bb{s}(\alpha) &= \bm{\mu}_s + \bb{U}_s\text{diag}(\bm{\sigma}_s)\bm{\alpha}, \\
    \bb{t}(\beta) &= \bm{\mu}_t + \bb{U}_t\text{diag}(\bm{\sigma}_t)\bm{\beta}.
\end{align}
$\bb{U}_\bullet$ contains the PCA directions and $\{\bm{\mu}_\bullet, \bm{\sigma}_\bullet\}$ are the PCA means and per-dimension standard deviations. We use our framework to reorient out-of-plane rotations and relight faces. This is a difficult task, because the encoder-decoder only sees 2D views of a self-occluding 3D scene. The encoder has to learn to decouple the complex interaction between light and 3D surfaces, while inferring missing information, then the decoder has to convert this representation into a faithful `rendering' of the transformed scene. A key difficulty is to infer occluded surfaces, which may be disoccluded upon out-of-plane rotation of the face.

We generate 1000 random RGB faces of size $150\times150$. For each face we generate 240 random views varying azimuth $[-43^\circ,43^\circ]$ and elevation $[-15^\circ,15^\circ]$ from head on (we use a right-handed coordinate frame with the $x$-axis pointing through the nose and $z$-axis pointing upwards), and with random lighting directions azimuth $[-57^\circ,57^\circ]$ and elevation $[-57^\circ,57^\circ]$. Both the rotations of the faces and lighting positions can be efficiently encoded using 3D rotation matrices $\bb{R}_{\text{rot}}$ and $\bb{R}_{\text{light}}$, each with 2 degrees of freedom---azimuth $\psi$ and elevation $\theta$, but no roll. Thus a natural form for the feature transform matrices, which we use is
\begin{align}
	\bb{F}_\theta &= \begin{bmatrix}
		\bb{R}_{\text{rot}} & \\
        & \bb{R}_{\text{light}} \\
	\end{bmatrix} \\
    \bb{R}_\bullet &= \begin{bmatrix}
		\cos \theta_\bullet 	&	& \sin \theta_\bullet 	\\
        						& 1	& 				\\
        -\sin \theta_\bullet	&	& \cos \theta_\bullet 	\\
	\end{bmatrix} \begin{bmatrix}
		\cos \psi_\bullet 	& -\sin \psi_\bullet 	&	\\
        -\sin \psi_\bullet	& \cos \psi_\bullet 	&	\\
        				&				& 1	\\
	\end{bmatrix}
\end{align}
We dub it the \emph{facial transformer}. As basic design principles, we avoid max-pooling, favoring strides, and use batch normalization and leaky ReLUs after all layers, apart from before and after the feature transform layer. For deconvolution we upsample with nearest-neighbor interpolation followed by regular convolution \cite{odena16,Kulkarni15}. Inspired by \cite{Xie16,Godard16} our reconstruction loss is a convex combination of the structural similarity index (SSIM) \cite{Wang04} and L1 loss. The L1 loss encourages low-frequency shape information and accurate color matching, and the SSIM encourages high-frequency details, for instance, the shading of the ears. The loss is
\begin{align}
	\mc{L}_{\text{face}} = \frac{\alpha}{N}\sum_{j\in\text{pixels}} \frac{1-\text{SSIM}(x_{j},\tilde{x}_{j})}{2} + (1-\alpha) \left | \bb{x}_j-\tilde{\bb{x}}_j \right |
\end{align}
where $N$ is number of pixels times 3 color channels. Similarly to \cite{Xie16,Godard16}, we use the blending coefficient of $\alpha=0.85$. We optimize the loss using Adam \cite{Kingma14}, minibatch size 32, and initial learning rate $10^{-4}$, dividing by 10 at iteration 30000 and 50000, for a total of 60000 iterations. We train on a single TITAN X Pascal GPU. \nicefrac{1}{4}-2 hours is sufficient for good results.
\begin{figure}[b]
\vspace{-8pt}
	\includegraphics[width=\linewidth]{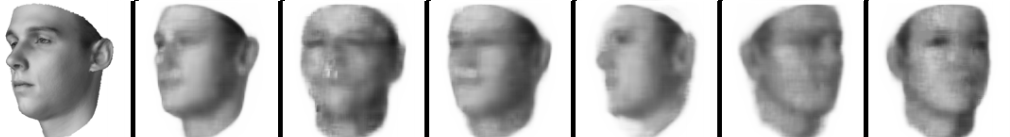}
    \includegraphics[width=\linewidth]{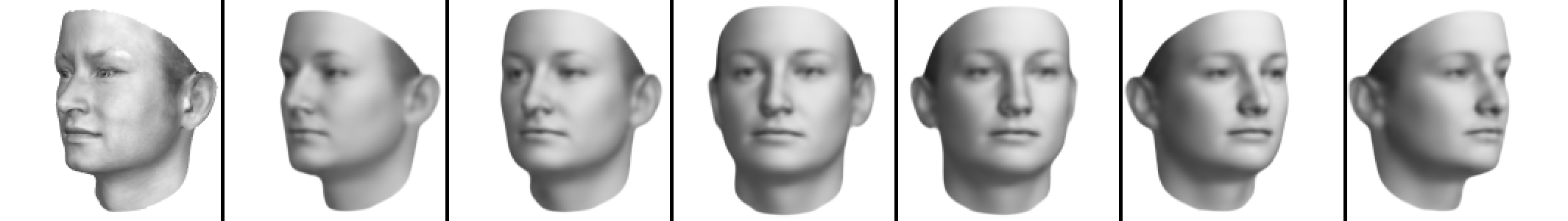}
    \caption{Side-by-side output of DC-IGN \cite{Kulkarni15} \textsc{top} and our facial transformer \textsc{bottom}. We have grayscaled our image for a fairer comparison. Input on left, smoothly rotated faces on right. We emphasize here that the goal of DC-IGN is different to ours, since they learn unsupervised disentangling. We argue to use supervision when the information is accessible. Our use of supervision is evident in that we can quantitatively rotate our faces; whereas, DC-IGN cannot.}
    \label{fig:dcign}
\end{figure}
Figure \ref{fig:faces_panel} shows the results of reoriented and relit faces from a held-out validation set. The input is on the left and the transformed outputs on the right. Top to bottom each row shows a different transformation, namely, lighting azimuth, lighting elevation, rotation azimuth, and rotation elevation. Faces inside the large green box span the transformation parameters seen at training time, those outside were not seen. We note the reconstruction fidelity and impressive ability to reorient out-of-plane rotations, but zooming in shows that the reconstructions lack high-frequency detail to be foolproof replicas of the input and the overall face shape changes slightly. For unseen transformation parameters, notice how faces just outside the green box are of similar quality to inside, but large deviations from the training set degrade. This is especially so for the geometric rotations, where the boundary surfaces (nose and chin in particular) begin to protrude from the face. Surprisingly, the shading of the faces is realistic outside of the box. We also compare against DC-IGN \cite{Kulkarni15} in Figure \ref{fig:dcign}. Our superior quality is partially down to better training, but also to improved alignment in feature-space, from supervised transformation information. Interpretability of our features allows for more accurate control over the azimuthal rotation.
\begin{figure}[b]
	\includegraphics[width=0.24\linewidth]{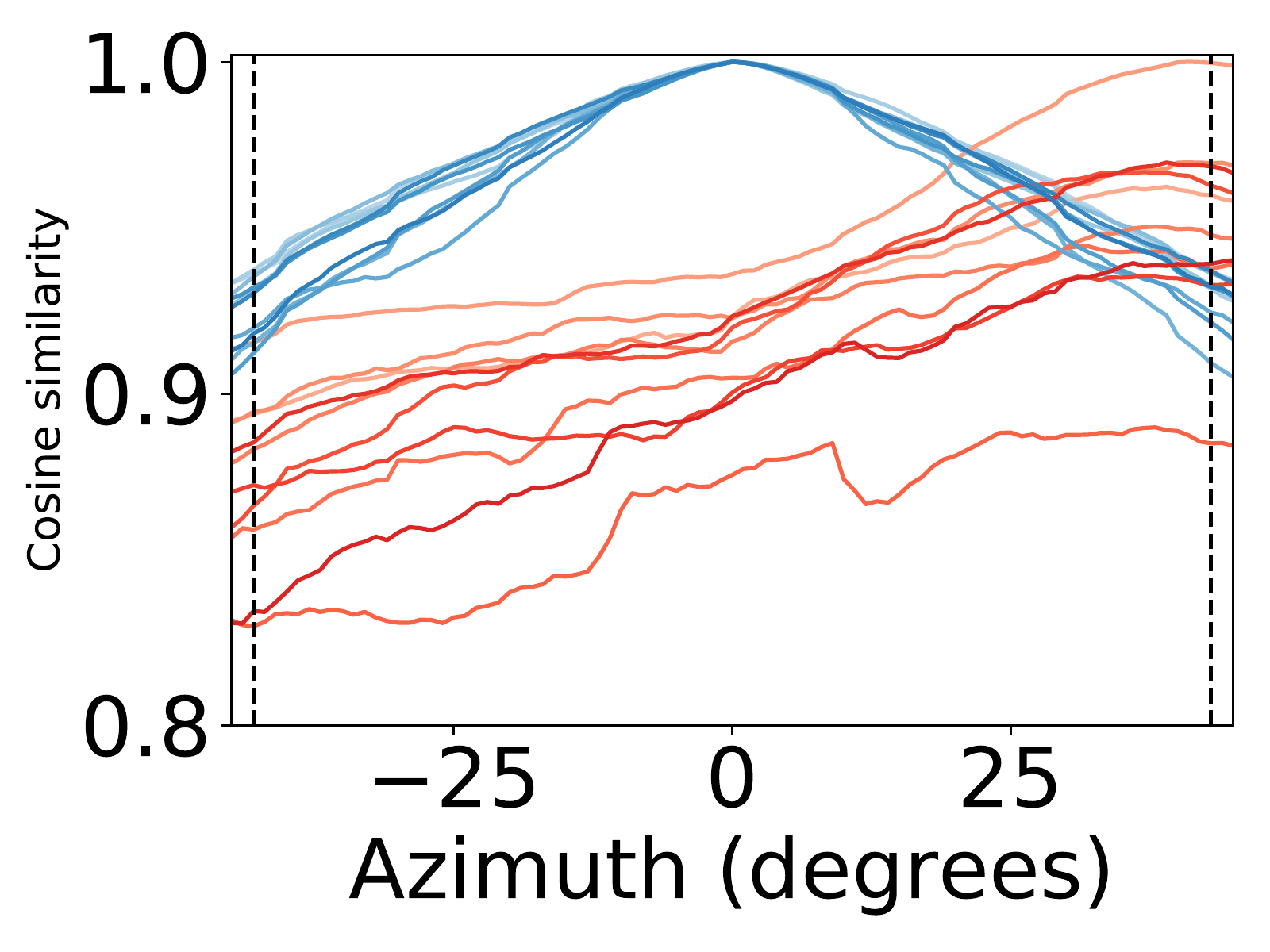}
    \includegraphics[width=0.24\linewidth]{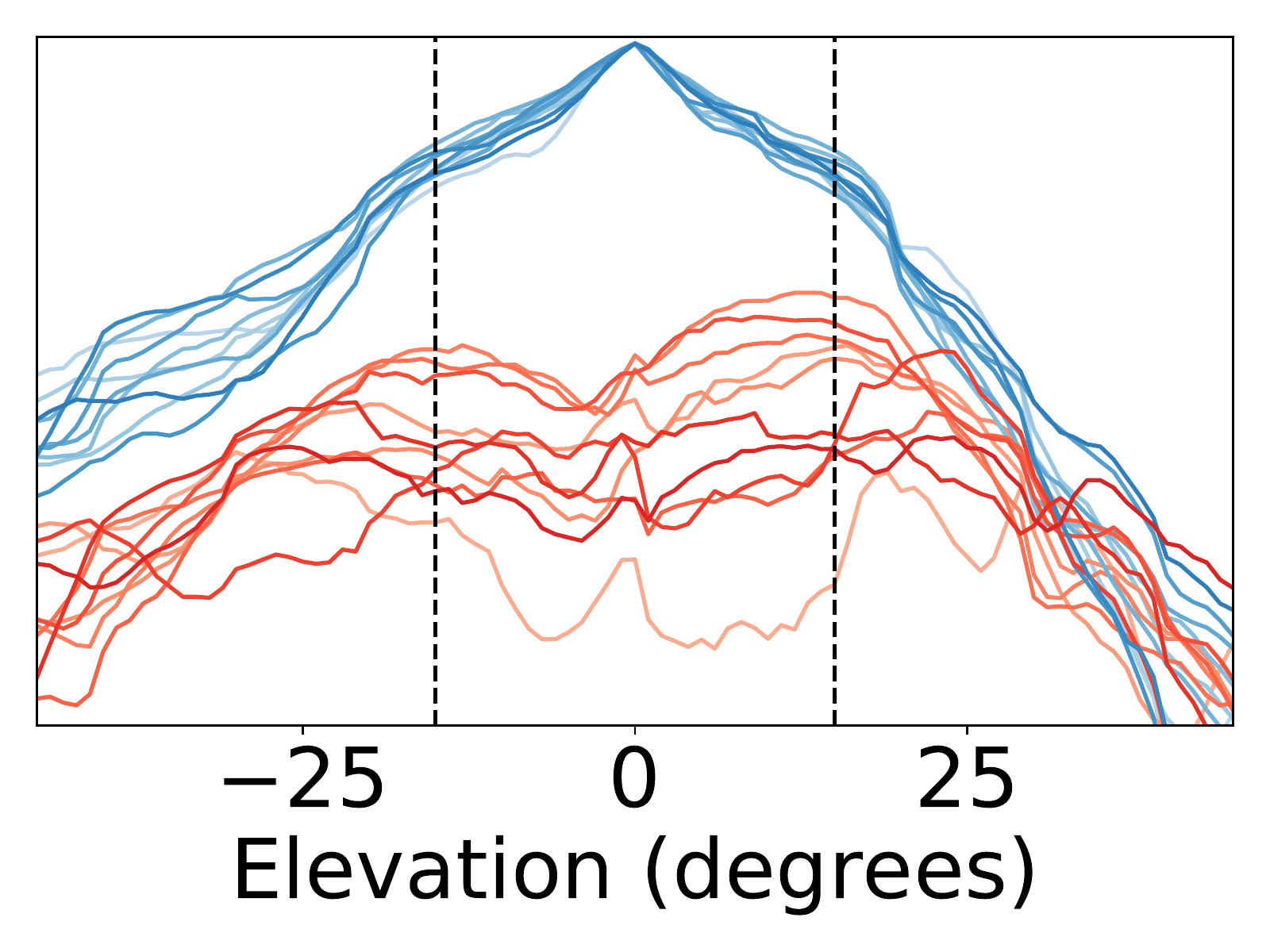}
    \includegraphics[width=0.24\linewidth]{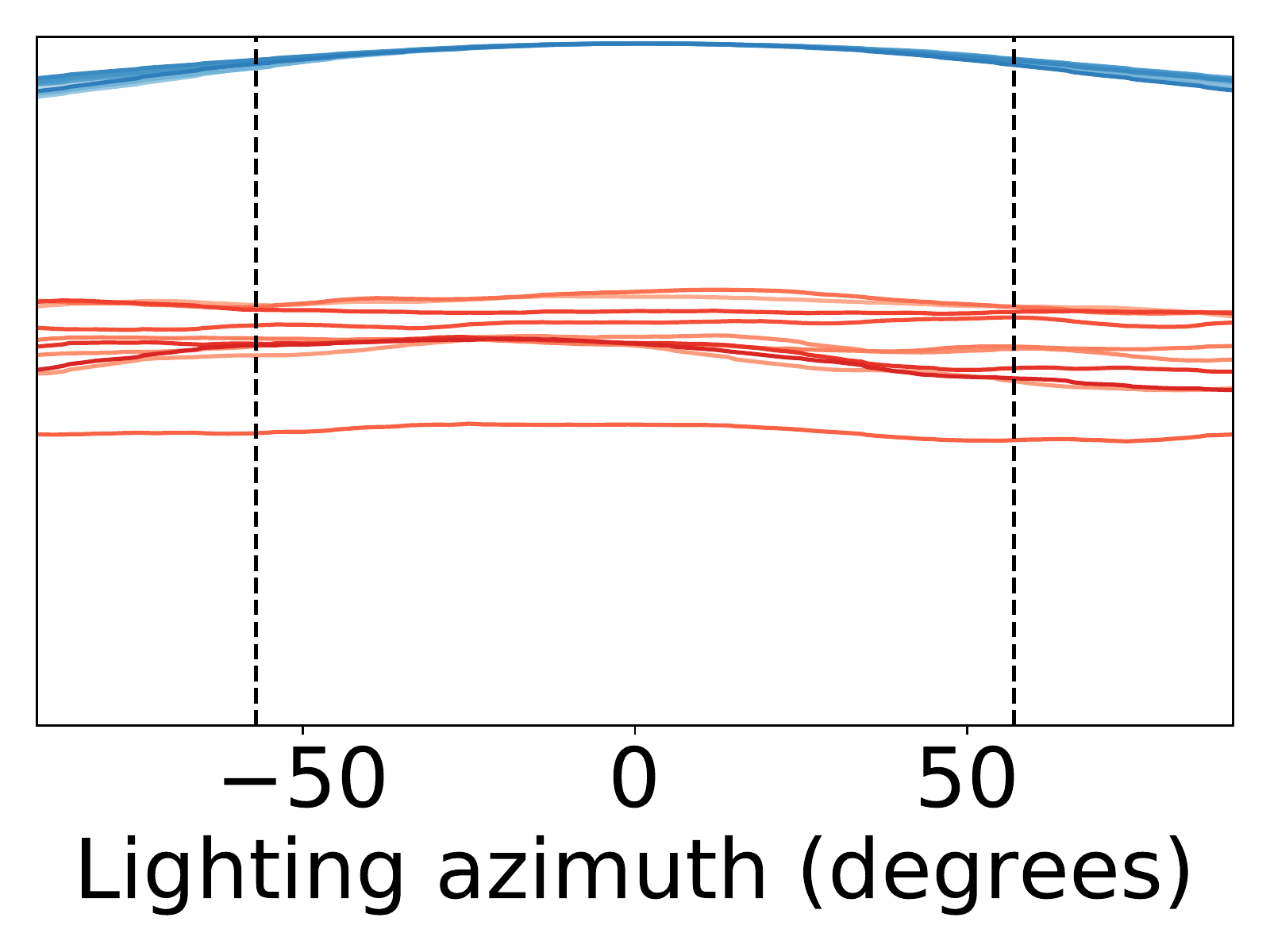}
    \includegraphics[width=0.24\linewidth]{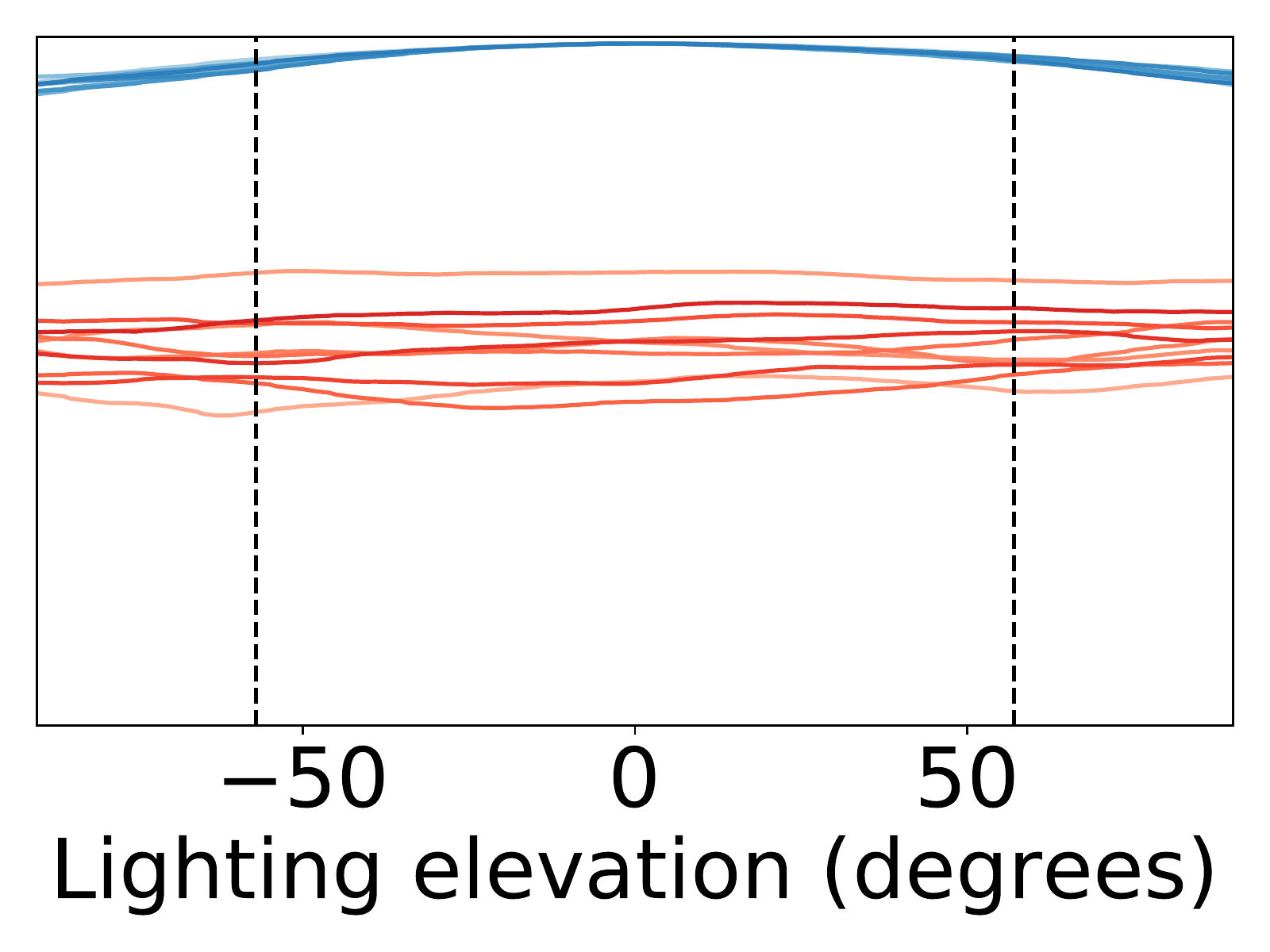}
    \includegraphics[width=0.24\linewidth]{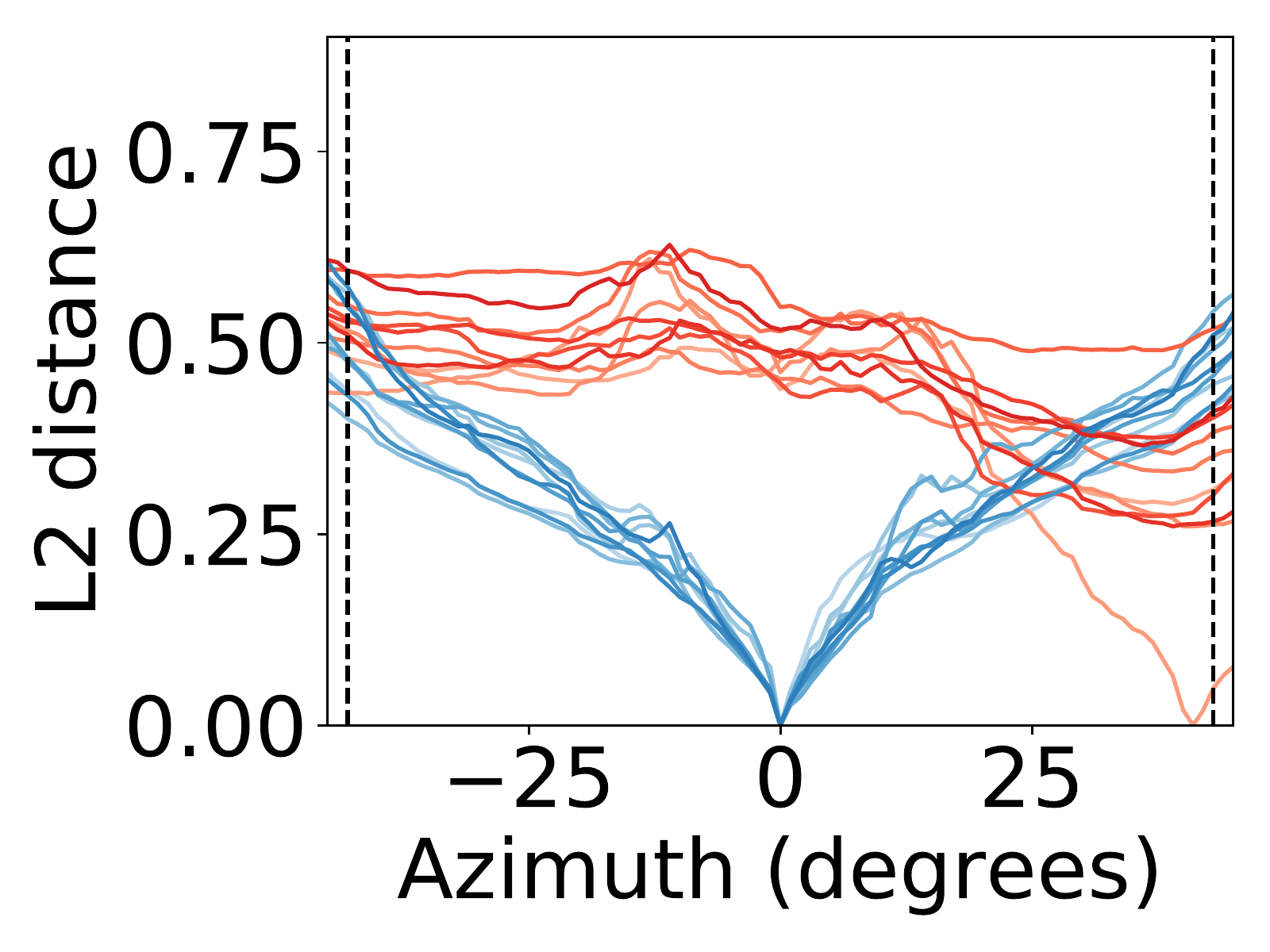}
    \includegraphics[width=0.24\linewidth]{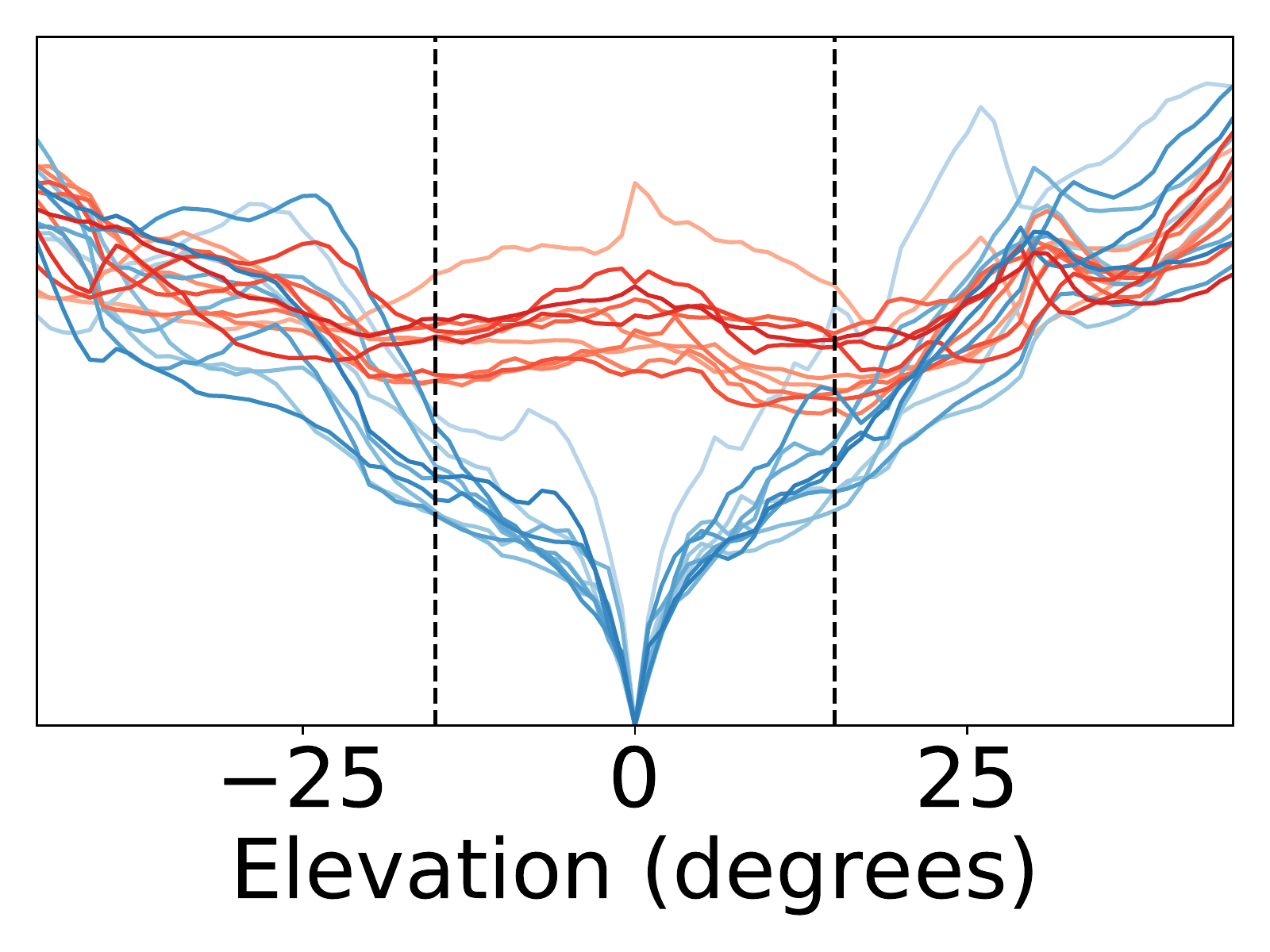}
    \includegraphics[width=0.24\linewidth]{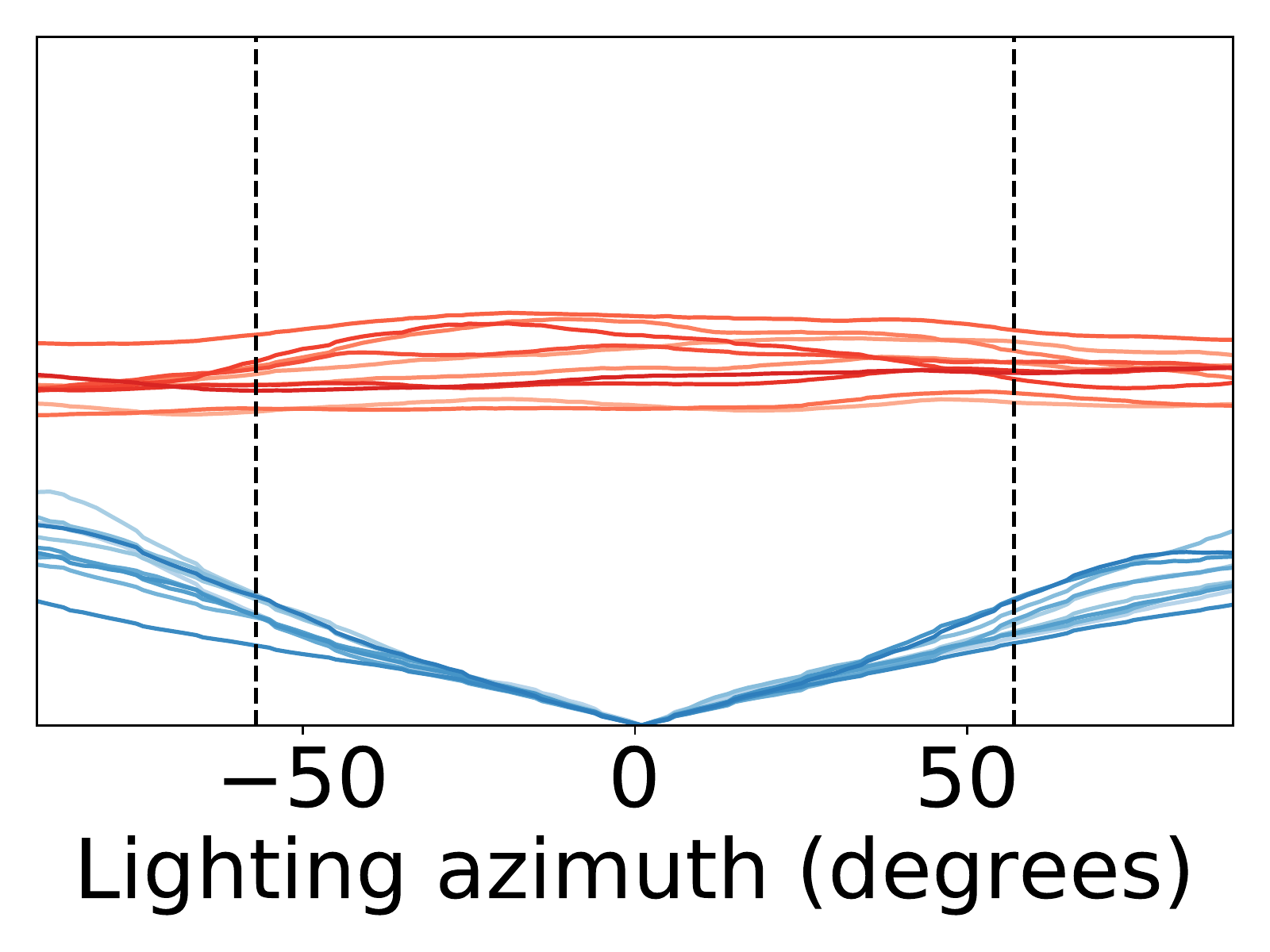}
    \includegraphics[width=0.24\linewidth]{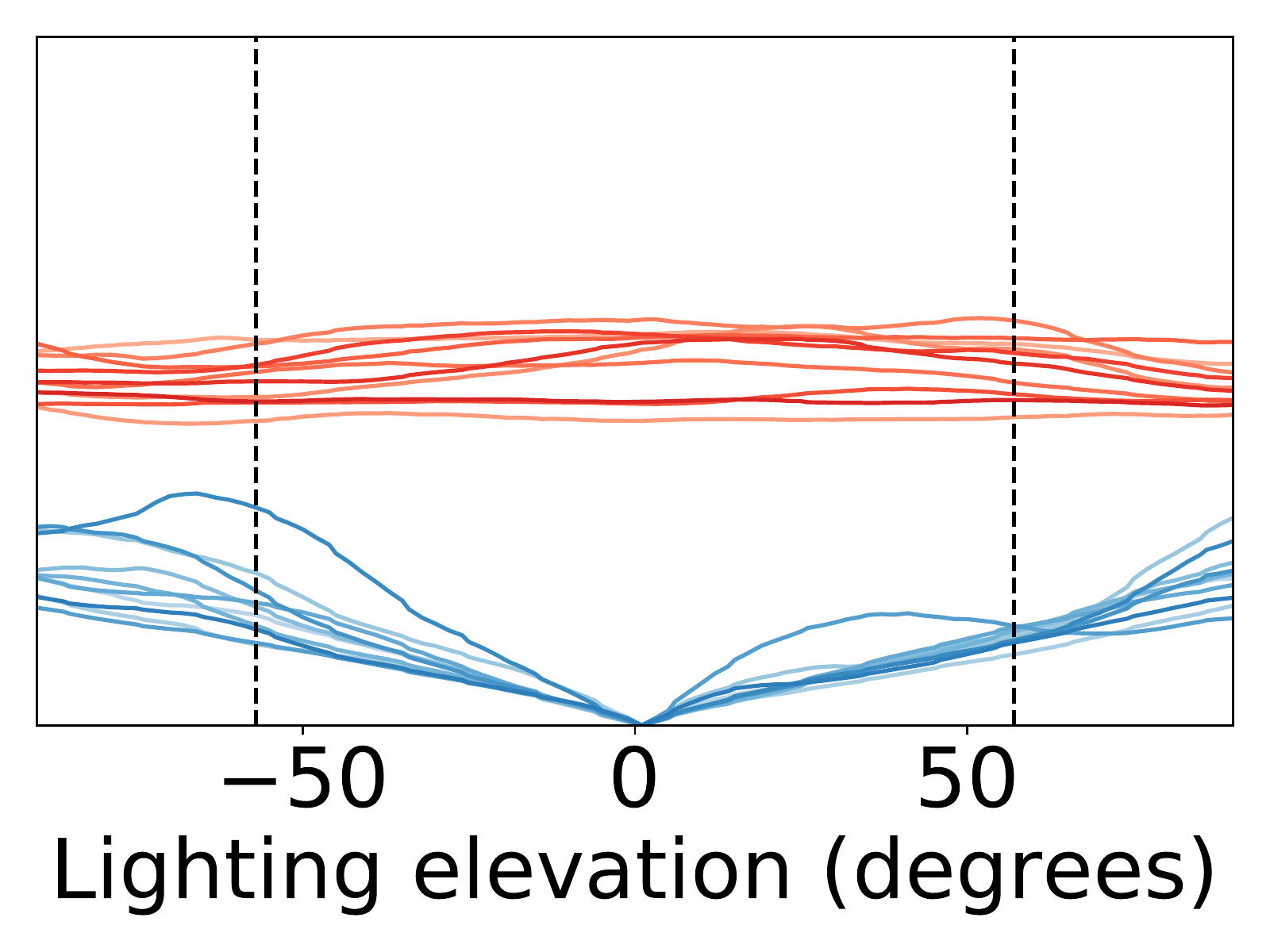}
    \caption{Pairs of images are compared to each other in feature space and similarity is measured using cosine similarity \textsc{top} and L2 distance \textsc{bottom}. Pairs of images with same identity shown in {\color{blue}blue}, and pairs with different identities shown in {\color{red}red} (10 each). Columns show left to right: sweeping of azimuth, elevation, lighting azimuth, and lighting elevation with all other parameters held. Dashed vertical lines show range of transformation values seen at training time. Ideally cosine similarity would be 1 everywhere for the blue lines, indicating perfect transformation invariance. For dissimilar faces, the red curves would be less than 1. We see that invariance to lighting is easy, even beyond the range of training examples (see central box in Figure \ref{fig:faces_panel}). Elevation is particularly hard, so two features of the same person begin to differ at large elevations.}
    \label{fig:stability}
\end{figure}
\begin{figure}[ht]
	\includegraphics[width=\linewidth]{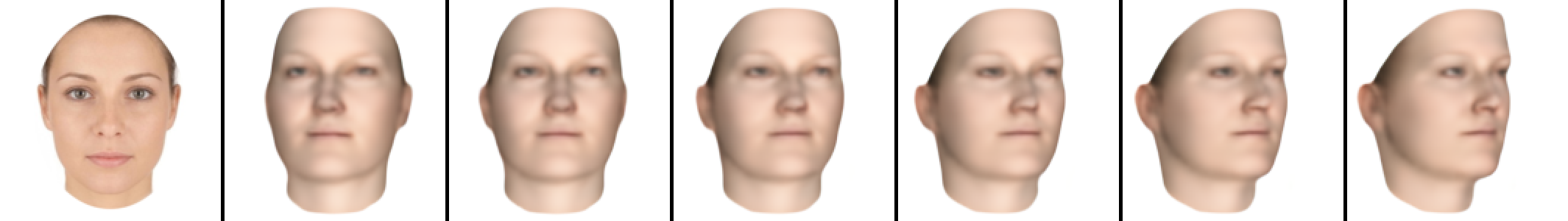}
    \includegraphics[width=\linewidth]{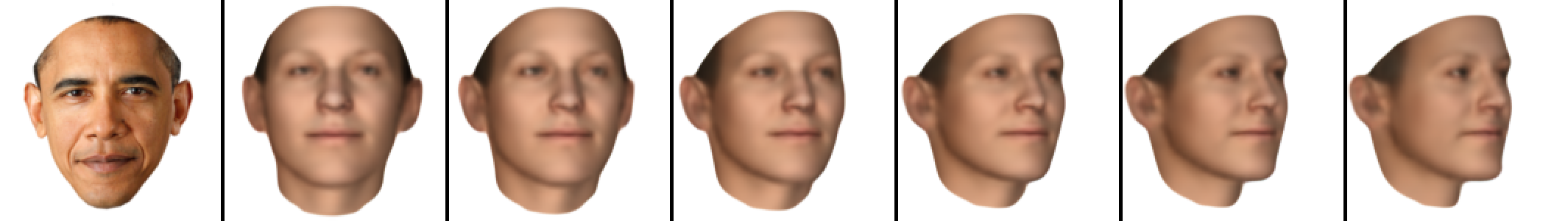}
    \includegraphics[width=\linewidth]{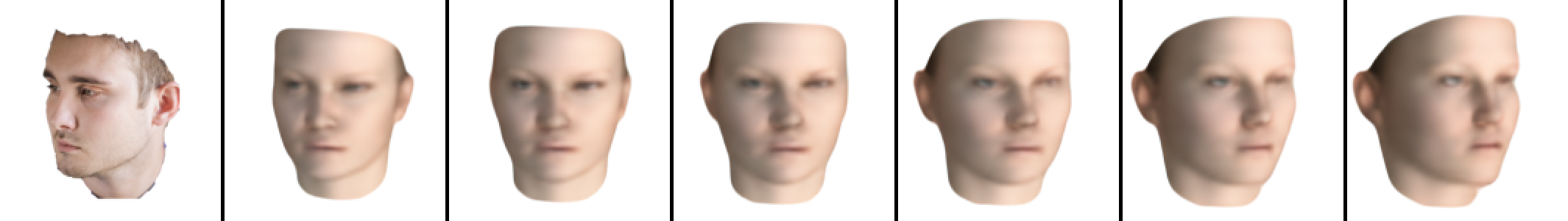}
    \caption{We pass images of real faces through our system re-orienting $50^\circ$ from the initial pose, while fixing all other transformation parameters. Despite being trained on artificial data, the system is able to extract basic pose, shape, appearance and illumination. The system struggles to match shape properly, since these are far from the training set.}
    \label{fig:real_faces}
\end{figure}
\begin{figure}[b]
\begin{center}	\includegraphics[width=0.90\linewidth]{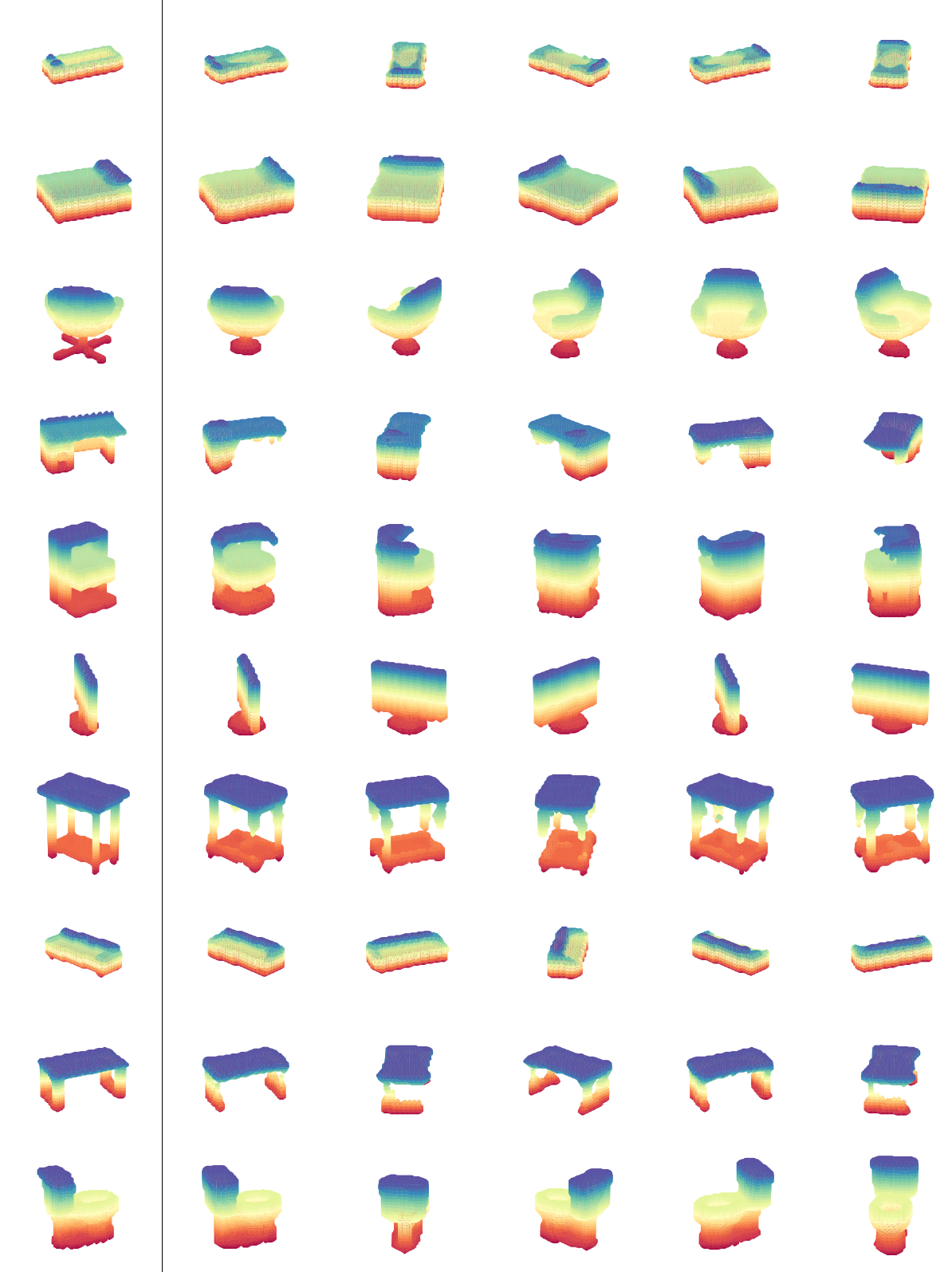}
    \caption{We pass randomly rotated volume of 10 categories from the test set (left) through our system re-orienting it by 0, 60, 120, 180, and 240 degrees from the initial pose. The system struggles to reconstruct thin shapes properly, which is a common problem due to sparseness of the volume occupancy.}
    \label{fig:shape_net}
\end{center}
\end{figure}
\textbf{Feature stability}
In Figure \ref{fig:stability} we test the feature stability under transformations of the input. We take an invariant representation of the data using L2-norms and relative phases, then measure the cosine similarity (top) and L2-distance (bottom) between a face and transformed versions of itself (blue), and we also compute these metrics between transformed versions of a face and a randomly selected face of another identity. There is a clear separation between faces of different identities for medium sized transformations, but this breaks down for large values of the parameters for geometric rotations. This is especially so, when the parameter values are close to the limit of the training data, as would be expected.

\textbf{Real faces}
For fun, we feed images of real faces into our system, to recognize basic pose, shape, appearance, and lighting. We take internet images, cropping out background and hair. The system makes crude, but convincing enough matches to pose, skin tone, and lighting. The bottom image is particularly hard due to the side pose and lighting. This shows our system has learned a generalizable representation of faces, despite training on artificial data.

%\subsection{Voxelized ShapeNets: 3D Volumetric Data --- 3D transformations}
\subsection{Voxelized ShapeNets:  3D Transformations}
For this experiment, we use the ModelNet10 subset of the ShapeNet dataset~\cite{ShapeNets15}. This consists of 3991 CAD models from 10 object categories. Specifically, we use the voxelized ModelNet10 provided by Maturana~\etal~\cite{maturana_iros_2015}, which is a volumetric binary occupancy grid of size 32x32x32.

The encoder-decoder architecture is similar to the variational auto-encoder architecture by Brock~\etal~\cite{brock2016generative}, with the bottleneck of 200 units with equivariance to rotations about the y-axis. We also employ their variant of the binary cross entropy loss for training:
\begin{align}
	\mc{L}_{\text{bce}} = \sum_{i\in\text{voxels}} -\gamma t_i \log(o_i) - (1-\gamma) (1-t_i) \log(1-o_i) \;,
\end{align}
where $t_i$ are the target values rescaled to $[-1,2]$, $o_i$ is the output of the auto-encoder rescaled to [0.1,0.9999] and $\gamma$ is set to $0.98$ to compensate for the sparseness of volumetric data. 
We optimize the loss using Adam, minibatch size 16, and learning rate of $10^{-4}$. See supplementary materials for details on classification.

\section{Conclusion}
We have presented a simple framework to learn deep feature-spaces, which disentangle both in-plane and out-of-plane transformations into an interpretable feature space, that also allows smooth interpolation. Our key innovation is the feature transform layer, which can be applied to both convolutional and fully-connected layers. The properties of the feature transform layer give our networks equivariance properties, that can help with generative and discriminative applications. %and allow it to be used for discriminative and fine control over feature-space transformations as well as to produce transformation invariants.

\textbf{Limitations}
Our approach is supervised, so labeled examples are needed to span the space of transformations, preferably with little other variety in the images. Also, the feature space needs to be smooth, precluding mirroring. 

\textbf{Acknowledgements}
Support is from Fight for Sight UK, a Microsoft Research PhD Scholarship, NERC NE/P016677/1, and NERC NE/P019013/1.

{\small
\bibliographystyle{ieee}
\bibliography{appendix}
}

\newpage
\onecolumn
\appendix

{\centering \large \textbf{Supplementary Material}}

%%%%%%%%% ABSTRACT
  \begin{abstract}
  Here we present ModelNet10 classification performance and mathematical definition of homomorphism property from the main paper.
  \end{abstract}

  \section{ShapeNets (ModelNet10) classification accuracy}
  The ModelNet10 classification task is evaluated on 908 models from the test set. For this task we trained the Modelnet architecture autoencoder with a 2-layer MLP (256-128-10) on the relative phase between all subvectors of the codes.

  We minimize the sum of two losses: cross-entropy loss for classification and the reconstruction loss. We follow~\cite{brock2016generative} for the binary cross-entropy reconstruction loss:
  \begin{align}
      \mc{L}_{\text{recon}} = \sum_{i\in\text{voxels}} -\gamma t_i \log(o_i) - (1-\gamma) (1-t_i) \log(1-o_i) \;,
  \end{align}
  where $t_i$ are the target values rescaled to $[-1,2]$, $o_i$ is the output of the autoencoder rescaled to [0.1,0.9999] and $\gamma$ is set to $0.98$ to compensate for the sparseness of volumetric data. 
  Thus, the loss is:
  \begin{align}
      \mc{L} = \mc{L}_{\text{recon}} + 10\mc{L}_{\text{classification}}
  \end{align}

  We optimize the loss using Adam and minibatch size 16, and learning rate of $10^{-4}$. We use the augmentation strategy of Maturana~\etal\cite{maturana_iros_2015}.

  We accurately classify 821 models out of 908, with an accuracy of 90.4\%.

  \begin{table}[h]
    \begin{center}
      \begin{tabular}{|l|c|}
      \hline
      Method & Accuracy \\
      \hline
      VRN Ensemble~\cite{brock2016generative}				&  97.14\% \\
      ORION~\cite{sedaghat2016orientation}									&  93.8\% \\
      LightNet~\cite{lightweight}									&  93.39\% \\
      FusionNet~\cite{hegde2016fusionnet}									&  93.11\% \\
      Pairwise~\cite{johns2016pairwise}									&  92.8\% \\
      GIFT~\cite{bai2016gift}									&  92.35\% \\
      VoxNet~\cite{maturana_iros_2015}
          &  92\% \\
      3D-GAN~\cite{wu2016learning}
          &  91.00\% \\

      Ours									&  90.4\% \\
      \hline
      \end{tabular}
    \end{center}
    \caption{State of the Art methods and their classification accuracy on ModelNet10 benchmark.}
    \label{tab:comparisons}
  \end{table}

  \section{The Homomorphism Property}
  The homomorphism property (Equation 6) is 
  \begin{align}
      \bb{F}_{\theta_2\theta_1} = \bb{F}_{\theta_2}\bb{F}_{\theta_1}.
  \end{align}
  Thus if $I\in\Theta$ is the identity transformation, then 
  \begin{align}
      \bb{F}_{\theta}=\bb{F}_{I\theta}=\bb{F}_{I}\bb{F}_{\theta} \implies \bb{F}_{I}=\bb{I},
  \end{align} 
  where $\bb{I}$ is the identity matrix. This in turn implies the invertability property $\bb{F}_{\theta^{-1}}=\bb{F}_{\theta}^{-1}$, since 
  \begin{align}
      \bb{I}=\bb{F}_{I}=\bb{F}_{\theta\theta^{-1}}=\bb{F}_{\theta}\bb{F}_{\theta^{-1}} \implies \bb{F}_{\theta^{-1}}=\bb{F}_{\theta}^{-1}.
  \end{align}

\end{document}